\newif\ifarXiv
\arXivtrue

\PassOptionsToPackage{sort}{natbib}

\documentclass[pmlr]{jmlr} 

\ifarXiv
    \newcommand{\tagdsmode}{nonproceedings}
\else%
    \newcommand{\tagdsmode}{proceedings}
\fi

\usepackage{etoolbox}
\robustify\bfseries

\makeatletter
\newcommand{\tagdssubmission}{submission}
\newcommand{\tagdsproceedings}{proceedings}

\ifx\tagdsmode\tagdsproceedings

\else\ifx\tagdsmode\tagdssubmission
  \def\ps@jmlrtps{%
    \let\@mkboth\@gobbletwo
    \def\@oddhead{\scriptsize Under Review at the 2nd Conference on Topology, Algebra, and Geometry in Data Science\hfill}%
    \let\@evenhead\@oddhead
    \def\@oddfoot{}%
    \let\@evenfoot\@oddfoot
  }

\else
  \def\ps@jmlrtps{%
    \let\@mkboth\@gobbletwo
    \def\@oddhead{}%
    \let\@evenhead\@oddhead
    \def\@oddfoot{}%
    \let\@evenfoot\@oddfoot
  }
\fi\fi
\makeatother

\usepackage{longtable}

\usepackage{booktabs}
\usepackage{siunitx}

\usepackage{paralist}

\theorembodyfont{\upshape}
\theoremheaderfont{\scshape}
\theorempostheader{:}
\theoremsep{\newline}

\jmlrvolume{334}
\jmlryear{2026}
\jmlrworkshop{Topology, Algebra, and Geometry in Data Science}

\title[Template]{TAG-DS Paper Template}

\ifx\tagdsmode\tagdssubmission

\else

  \author{\Name{Nello Blaser} \Email{nello.blaser@uib.no} \\
   \Name{Odin Hoff Gardaa} \Email{odin.garda@uib.no} \\
   \Name{Lars M. Salbu} \Email{lars.salbu@uib.no} \\
   \addr Department of Informatics, University of Bergen, Norway \\
   \AND
   \Name{Elena Xinyi Wang} \Email{xinyi.wang@unifr.ch} \\
   \Name{Bastian Rieck} \Email{bastian.grossenbacher@unifr.ch}\\
   \addr Department of Informatics, University of Fribourg, Switzerland}

\fi

\usepackage{graphicx} 
\usepackage[table]{xcolor} 
\usepackage{multirow, makecell}
\usepackage[utf8]{inputenc}
\usepackage{hyperref}
\usepackage{tikz}
\usetikzlibrary{positioning, arrows.meta, fit, calc, matrix, decorations.pathreplacing, backgrounds}
\usepackage{float}
\usepackage{tcolorbox}
\usepackage[capitalize, noabbrev]{cleveref}
\usepackage{microtype}
\usepackage{enumitem}
\usepackage[american]{babel}
\usepackage{silence}
\ifarXiv
    \newcommand{\repo}{https://github.com/blasern/ect-representation}
\else
    \newcommand{\repo}{https://anonymous.4open.science/r/ect-representation-CAC2}
\fi

 \hypersetup{
     colorlinks=true,
     linkcolor=cyan,
     filecolor=cyan,
     citecolor=cyan, 
     urlcolor=cyan,
     }

\tcbset{
  boxsep     = 1.0mm,
  colframe   = cyan,
  colback    = cyan!10,
  left       = 1.0mm,
  right      = 1.0mm,
}

\usetikzlibrary{arrows.meta, positioning, fit, backgrounds}

\definecolor{datacol}{RGB}{220,235,252}
\definecolor{layercol}{RGB}{255,224,189}
\definecolor{choicecol}{RGB}{209,236,214}
\definecolor{bordercol}{RGB}{60,60,60}
\definecolor{veccol}{RGB}{90,140,220}

\tikzset{
  datanode/.style={
    rectangle, rounded corners=4pt,
    draw=bordercol, line width=0.7pt,
    fill=datacol, inner sep=5pt,
    font=\footnotesize\sffamily, align=center
  },
  layernode/.style={
    rectangle, rounded corners=3pt,
    draw=bordercol, line width=0.7pt,
    fill=layercol, minimum width=2.2cm, minimum height=1.0cm,
    font=\small\sffamily, align=center
  },
  choicenode/.style={
    rectangle, rounded corners=3pt,
    draw=bordercol, line width=0.7pt,
    fill=choicecol, minimum width=2.7cm, minimum height=0.72cm,
    font=\small\sffamily, align=center
  },
  arrow/.style={-Stealth, thick, color=bordercol},
  groupbox/.style={
    draw=bordercol, dashed, rounded corners=5pt,
    line width=0.6pt, fill=choicecol!25, inner sep=7pt
  }
}

\newcommand{\drawgraph}{%
  \begin{tikzpicture}[scale=0.60, baseline=0.35cm]
    \draw[gray!60, line width=0.9pt] (0,0.4)--(0.7,1.1);
    \draw[gray!60, line width=0.9pt] (0,0.4)--(0.9,0.0);
    \draw[gray!60, line width=0.9pt] (0.7,1.1)--(1.6,0.8);
    \draw[gray!60, line width=0.9pt] (0.9,0.0)--(1.6,0.8);
    \draw[gray!60, line width=0.9pt] (0.7,1.1)--(0.9,0.0);
    \draw[gray!60, line width=0.9pt] (0.9,0.0)--(1.5,-0.5);
    \draw[gray!60, line width=0.9pt] (0,0.4)--(0.2,-0.4);
    \foreach \px/\py in {0/0.4, 0.7/1.1, 0.9/0.0, 1.6/0.8, 1.5/-0.5, 0.2/-0.4}
      \filldraw[fill=white, draw=bordercol, line width=0.6pt] (\px,\py) circle (0.12);
  \end{tikzpicture}%
}

\newcommand{\drawect}{%
  \begin{tikzpicture}[baseline=0.75cm]
    \foreach \r/\c/\v in {%
      0/0/0, 0/1/0, 0/2/1, 0/3/1, 0/4/2, 0/5/2, 0/6/1, 0/7/0,%
      1/0/0, 1/1/1, 1/2/1, 1/3/2, 1/4/2, 1/5/1, 1/6/0, 1/7/0,%
      2/0/0, 2/1/0, 2/2/0, 2/3/1, 2/4/2, 2/5/2, 2/6/1, 2/7/0,%
      3/0/0, 3/1/1, 3/2/2, 3/3/2, 3/4/2, 3/5/1, 3/6/1, 3/7/0,%
      4/0/0, 4/1/0, 4/2/1, 4/3/1, 4/4/1, 4/5/2, 4/6/1, 4/7/0,%
      5/0/0, 5/1/1, 5/2/1, 5/3/2, 5/4/1, 5/5/1, 5/6/0, 5/7/0%
    }{%
      \ifcase\v\relax
        \fill[white]     (\c*0.27,\r*0.27) rectangle ++(0.27,0.27);
      \or
        \fill[veccol!45] (\c*0.27,\r*0.27) rectangle ++(0.27,0.27);
      \or
        \fill[veccol!90] (\c*0.27,\r*0.27) rectangle ++(0.27,0.27);
      \fi
      \draw[gray!35, line width=0.2pt] (\c*0.27,\r*0.27) rectangle ++(0.27,0.27);
    }
    \draw[bordercol, line width=0.5pt] (0,0) rectangle (2.16,1.62);
    \node[font=\tiny\sffamily, rotate=90, anchor=south] at (-0.12, 0.81) {$D$};
    \node[font=\tiny\sffamily, anchor=north] at (1.08, -0.08) {$H$};
  \end{tikzpicture}%
}

\newcommand{\drawzvec}{%
  \begin{tikzpicture}[baseline=1.2cm]
    \foreach \i/\sh in {%
      0/55, 1/30, 2/80, 3/45, 4/70, 5/90,%
      6/35, 7/60, 8/25, 9/75, 10/50, 11/40%
    }{%
      \fill[veccol!\sh] (0,\i*0.22) rectangle (0.50,\i*0.22+0.20);
      \draw[gray!35, line width=0.2pt] (0,\i*0.22) rectangle (0.50,\i*0.22+0.20);
    }
    \draw[bordercol, line width=0.5pt] (0,0) rectangle (0.50,2.64);
  \end{tikzpicture}%
}

\newcommand{\drawyout}{%
  \begin{tikzpicture}[baseline=0.2cm]
    \fill[veccol!65] (0,0) rectangle (0.52,0.52);
    \draw[bordercol, line width=0.6pt] (0,0) rectangle (0.52,0.52);
  \end{tikzpicture}%
}

\newcommand{\drawtokens}{%
  \begin{tikzpicture}[scale=0.18, baseline=0pt]
    \foreach \y in {0,1,2} {
      \draw[bordercol, fill=bordercol!10] (0, \y) rectangle ++(1, 1);
      \draw[bordercol, fill=bordercol!10] (1, \y) rectangle ++(1, 1);
    }
  \end{tikzpicture}%
}

\title{Encoding the Euler Characteristic Transform}
\date{}

\begin{document}

\maketitle
\begin{abstract}
    The Euler Characteristic Curve (ECC) records the Euler characteristic of a linearly embedded cell complex as a function of filtration height in a given direction, and the Euler Characteristic Transform (ECT) is the injective shape descriptor obtained by collecting ECCs over many directions.
    How the ECT is encoded for a neural network is itself an inductive bias, conventionally fixed by discretizing each ECC.
    We introduce a continuous encoding: for each direction and each vertex it records the net Euler-characteristic change attributed to that vertex, producing a per-direction token sequence that a small transformer maps to a feature vector.
    We separate the resulting pipeline into two stages on orthogonal axes: an \emph{ECC encoder} that acts within each direction, mapping its curve to a fixed-length vector, and an \emph{ECT representation} that acts across directions, aggregating the per-direction vectors into one. 
    We study six ECT representation architectures spanning a range of inductive biases, from a structure-agnostic feedforward baseline to convolutional and complex-valued models that preserve equivariance under planar rotations.
    Across six classification benchmarks covering point clouds, graphs, cubical complexes, and meshes, the continuous encoding improves accuracy on all six datasets, and control experiments attribute the gain to the tokenization itself rather than to the added transformer capacity. 
    The representation architecture matters less than the encoding, and the payoff from its inductive biases depends on the encoding: a feedforward network performs best under continuous encoding but is less robust under discretization than convolutional architectures.
\end{abstract}

\section{Introduction}
The Euler Characteristic Transform (ECT) has emerged as a powerful descriptor for shape analysis, combining topological connectivity information and geometric structure into a computationally efficient representation. Originally introduced by \citet{turner2014persistent}, the ECT operates by filtering a shape along multiple directions and computing the Euler characteristic---a topological invariant that captures information about connected components, holes, and voids---at each filtration step. The resulting transform has been proven injective for embedded simplicial complexes in $\mathbb{R}^d$ \citep{MR1448182,turner2014persistent,Ghrist2018,curry2022many}, making it a sufficient descriptor for shape comparison and classification tasks. 
The Differentiable Euler Characteristic Transform (DECT) of \citet{roell2024differentiable} made the computation differentiable with respect to both sampling directions and input coordinates, enabling end-to-end learning within neural networks.
The (D)ECT has subsequently been used in several distinct roles in neural network pipelines:
\begin{inparaenum}[(i)]
    \item as a global shape descriptor fed to a classifier \citep{roell2024differentiable, mcguire2024classification, toscano2025molecular},
    \item as a local feature for graph representation learning \citep{rohrscheidt2025disslect, amboage2026leap},
    \item as a topology-aware loss term \citep{nadimpalli2023euler, roell2024differentiable}, and \item as an invertible descriptor for generative modeling \citep{roell2026ipt}.
\end{inparaenum}
Our work is squarely in the first category and asks how the global ECT should be encoded once it has been produced.

The integration of ECT into neural network architectures has taken several distinct approaches, with different strategies for encoding the directional and multi-scale information inherent in the transform. In existing work the ECT is typically discretized first, i.e., the Euler Characteristic Curve (ECC) along each direction is evaluated at a fixed grid of height thresholds, producing a fixed-dimensional matrix that the downstream network operates on. Prior architectures process this matrix either as an unordered set of curves, via a DeepSets \citep{zaheer2017deep} model \citep{roell2024differentiable}, or as a 2D image fed to a standard CNN \citep{mcguire2024classification, roell2024differentiable}.

The encoding strategies described above all act on a discretized ECT: each ECC is evaluated at a fixed grid of $H$ height thresholds, producing a $D \times H$ matrix, where $D$ is the number of directions. The grid is a modeling choice with real costs. The resolution $H$ becomes a hyperparameter that has to be tuned per dataset, trading expressivity against memory and compute---and the natural scale of that trade-off varies between datasets, since the relevant features in a $28 \times 28$ silhouette and in a high-resolution building mesh ``live'' at very different scales. The grid also forces the same resolution in regions where the ECC is changing rapidly and in regions where it is flat, even though all of the information can be attributed to the vertices. This suggests an alternative: encode the ECT by the steps \emph{themselves} rather than by samples of the function between them. A representation that records what changes, where, and by how much immediately sidesteps the resolution hyperparameter entirely and is exact by construction.

\paragraph{Contributions.} This paper makes the following contributions:
\begin{itemize}
    \item We introduce a \emph{continuous} representation of the ECT that records, for each sampling direction and each vertex of the complex, the net Euler characteristic change attributed to that vertex. The representation is grid-free, exact, parsimonious, and inherently differentiable, and a small transformer encoder maps it to a per-direction feature vector that replaces the discretized ECT. 
    \item We systematically study six neural network architectures for representing the ECT: a feedforward network, Deep Set, 1D convolution, 2D convolution, 1D complex convolution, and a hybrid model. The latter four are designed to preserve equivariance under planar rotations of the input; the first two serve as baselines without this structure.
    \item We provide an empirical evaluation across six classification benchmarks for different data types---point clouds, graphs, cubical complexes, and meshes---comparing all architectures with both continuous and discrete ECT representations, with attention to both accuracy and training stability.
\end{itemize}

\section{Continuous ECC Encoding} \label{sec:continuous}

We propose a novel \emph{continuous} representation that stores the ECCs exactly, without using a grid. For the sake of notational simplicity, we reserve our exposition to the 2D case. For a cell complex $K$ linearly embedded in $\mathbb{R}^2$ with $N$ vertices, and a given direction $\mathbf{w}_i = (\cos\theta_i, \sin\theta_i)$ corresponding to angle $\theta_i$, the filtration assigns to each cell $\tau \in K$ the height
\begin{equation*}
    h(\tau, \mathbf{w}_i)
    = \max_{v \in \tau} \langle v, \mathbf{w}_i \rangle, 
\end{equation*}
which is attained at the single vertex of $\tau$ with maximum projection onto $\mathbf{w}_i$. Define the \emph{responsible vertex} of a cell $\tau$ in direction $\mathbf{w}_i$ as
\begin{equation*}
    r(\tau, \mathbf{w}_i) = \arg\max_{v \in \tau} \langle v, \mathbf{w}_i \rangle.
\end{equation*}
Each cell can then be attributed entirely to its responsible vertex. For each vertex $v$ and direction $\mathbf{w}_i$, the net change in Euler characteristic as the filtration sweeps through $\langle v, \mathbf{w}_i \rangle$ is
\begin{equation*}
    \Delta\chi(v, \mathbf{w}_i) \;=\; \sum_{k \ge 0} (-1)^k \bigl|\{\, \tau \in \text{$k$-cells} : r(\tau, \mathbf{w}_i) = v \,\}\bigr|,
\end{equation*}
combining the vertex's own contribution with the signed contributions of every higher-dimensional cell that enters the filtration at the same threshold. Summing $\Delta\chi(v, \mathbf{w}_i)\,\mathbf{1}[h \ge \langle v, \mathbf{w}_i \rangle]$ over all vertices reproduces the ECC exactly, and the discretized ECT can be recovered at any resolution. 
Vertices with $\Delta\chi(v, \mathbf{w}_i) = 0$ for all directions $(\mathbf{w}_i)$ carry no information and are discarded, leaving $N_{\mathrm{active}} \le N$ tokens. For point clouds, where the complex has no higher-dimensional cells, $\Delta\chi \equiv 1$ at every vertex and $N_{\mathrm{active}} = N$.
Each vertex contributes its height $\langle v, \mathbf{w}_i \rangle$ as a real-valued token coordinate that is linear in the vertex position and smooth in the sampling angle, so the representation is differentiable by construction. The multiplicities $\Delta\chi(v, \mathbf{w}_i)$ are integer-valued and piecewise constant in the vertex positions, so they carry no gradient and are non-smooth only on a measure-zero set of tied projections.

To learn directly from the continuous representation, we process a sequence of tokens for each direction with a shared transformer encoder. For a given (cell complex, direction) pair, the input is a sequence of 2-dimensional tokens $\{(\langle v, \mathbf{w}_i \rangle, \Delta\chi(v, \mathbf{w}_i))\}_v$. This tokenization is also shown in \cref{fig:ecc}. A linear input projection maps each token to a $d_{\mathrm{model}}$-dimensional embedding, and a transformer encoder exchanges information across tokens within the same direction. Max-pooling over the token axis aggregates the variable-length sequence into a fixed-size vector per (cell complex, direction) pair, and a final linear projection produces a $d_{\mathrm{out}}$-dimensional output. The result is a $D \times d_{\mathrm{out}}$-dimensional matrix that takes the place of the discretized ECT matrix as input to any of the neural representation architectures described in \cref{sec:ect-representation}. Within a single cell complex, all $D$ directions are batched together for the transformer; cell complexes are processed sequentially to avoid padding to a common token count across the batch.

\definecolor{datacolor}{RGB}{210, 226, 246}
\definecolor{datacolorD}{RGB}{80, 125, 185}
\definecolor{opcolor}{RGB}{255, 224, 184}
\definecolor{opcolorD}{RGB}{210, 145, 60}
\definecolor{accentcolor}{RGB}{45, 70, 130}
\definecolor{notecolor}{RGB}{105, 105, 115}
\definecolor{activecolor}{RGB}{200, 70, 50}
\definecolor{inactivecolor}{RGB}{150, 150, 160}
\definecolor{insetcolor}{RGB}{248, 248, 252}
\definecolor{insetborder}{RGB}{200, 200, 215}

\begin{figure}
\resizebox{\textwidth}{!}{%
\begin{tikzpicture}[
    >=Latex,
    font=\sffamily,
    vdot/.style={circle, fill=black, inner sep=0pt, minimum size=3.6pt},
    activevdot/.style={circle, fill=activecolor, draw=activecolor!50!black,
                       line width=0.4pt, inner sep=0pt, minimum size=4.8pt},
    paneltitle/.style={font=\bfseries\small, text=accentcolor, anchor=south},
]
 
\def\sc{0.95}
\def\ystep{0.85}
 
\begin{scope}[local bounding box=panelA]
 
  \node[paneltitle] at (0, 2.0) {(a) Embedded complex $K$};
 
  \coordinate (v1) at (-2.5*\sc,  0.0);
  \coordinate (v2) at (-1.5*\sc,  0.8);
  \coordinate (v3) at (-1.0*\sc, -0.8);
  \coordinate (v4) at ( 0.5*\sc,  0.8);
  \coordinate (v5) at ( 1.0*\sc, -0.8);
  \coordinate (v6) at ( 2.5*\sc,  0.0);
 
  \fill[black!25, draw=black, thick] (v1) -- (v2) -- (v3) -- cycle;
   
  \draw[thick] (v2) -- (v4);
  \draw[thick] (v3) -- (v5);
  \draw[thick] (v4) -- (v5);
  \draw[thick] (v4) -- (v6);
  \draw[thick] (v5) -- (v6);
 
  \node[activevdot] at (v1) {};
  \node[activevdot] at (v5) {};
  \node[activevdot] at (v6) {};
  \node[vdot] at (v2) {};
  \node[vdot] at (v3) {};
  \node[vdot] at (v4) {};
 
  \node[font=\small, left=1.5mm]  at (v1) {$v_1$};
  \node[font=\small, above=1mm]   at (v2) {$v_2$};
  \node[font=\small, below=1mm]   at (v3) {$v_3$};
  \node[font=\small, above=1mm]   at (v4) {$v_4$};
  \node[font=\small, below=1mm]   at (v5) {$v_5$};
  \node[font=\small, right=1.5mm] at (v6) {$v_6$};
 
  \draw[->, very thick, accentcolor] (-2.3*\sc, -1.55) -- (-1.3*\sc, -1.55);
  \node[font=\small, text=accentcolor, anchor=west]
      at (-1.25*\sc, -1.55) {$\mathbf{w}_i = (1, 0)$};
 
\end{scope}
 
\begin{scope}[xshift=8.0cm, local bounding box=panelB]
 
  \node[paneltitle, align=center, text width=8cm] at (0, 2.0)
      {(b) Euler Characteristic Curve};
 
  \draw[->, thin] (-3.15*\sc, 0) -- (3.2*\sc, 0)
      node[right, font=\small] {$h$};
  \draw[->, thin] (-3.15*\sc, -1.4*\ystep - 0.1) -- (-3.15*\sc, 1.5*\ystep + 0.3)
      node[above, font=\small] {$\chi_h$};
 
  \foreach \y in {-1, 0, 1} {
    \draw[thin] (-3.2*\sc, \y*\ystep) -- (-3.1*\sc, \y*\ystep);
    \node[font=\scriptsize, left=1.5mm] at (-3.15*\sc, \y*\ystep) {$\y$};
  }
 
  \foreach \h in {-2.5, -1.5, -1, 0.5, 1, 2.5} {
    \draw[thin] (\h*\sc, 0.06) -- (\h*\sc, -0.06);
  }
 
  \draw[very thick, activecolor] (-3.1*\sc, 0) -- (-2.5*\sc, 0);
  \draw[very thick, activecolor] (-2.5*\sc, 0) -- (-2.5*\sc, 1*\ystep);
  \draw[very thick, activecolor] (-2.5*\sc, 1*\ystep) -- (1.0*\sc, 1*\ystep);
  \draw[very thick, activecolor] (1.0*\sc, 1*\ystep) -- (1.0*\sc, 0);
  \draw[very thick, activecolor] (1.0*\sc, 0) -- (2.5*\sc, 0);
  \draw[very thick, activecolor] (2.5*\sc, 0) -- (2.5*\sc, -1*\ystep);
  \draw[very thick, activecolor] (2.5*\sc, -1*\ystep) -- (3.15*\sc, -1*\ystep);
 
  \foreach \h/\v/\active in {-2.5/v_1/1, -1.5/v_2/0, -1/v_3/0, 0.5/v_4/0, 1/v_5/1, 2.5/v_6/1} {
    \node[font=\scriptsize, below=2mm, fill=white, inner sep=0.3mm]
        at (\h*\sc, -0.06) {$\h$};
    \ifnum\active=1
      \node[font=\scriptsize, text=activecolor, below=5.5mm,
            fill=white, inner sep=0.3mm] at (\h*\sc, -0.06) {$\v$};
    \else
      \node[font=\scriptsize, text=inactivecolor!50!black, below=5.5mm,
            fill=white, inner sep=0.3mm] at (\h*\sc, -0.06) {$\v$};
    \fi
  }
 
  \fill[activecolor] (-2.5*\sc, 1*\ystep) circle (2pt);
  \draw[thick, activecolor, fill=white] (-2.5*\sc, 0)         circle (2pt);
  \fill[activecolor] ( 1.0*\sc, 0)        circle (2pt);
  \draw[thick, activecolor, fill=white] ( 1.0*\sc, 1*\ystep)  circle (2pt);
  \fill[activecolor] ( 2.5*\sc, -1*\ystep) circle (2pt);
  \draw[thick, activecolor, fill=white] ( 2.5*\sc, 0)         circle (2pt);
 
  \node[font=\small\bfseries, text=activecolor, anchor=west]
      at (-2.5*\sc + 0.12, 0.5*\ystep) {$+1$};
  \node[font=\small\bfseries, text=activecolor, anchor=east]
      at ( 1.0*\sc - 0.12, 0.5*\ystep) {$-1$};
  \node[font=\small\bfseries, text=activecolor, anchor=west]
      at ( 2.5*\sc + 0.12, -0.5*\ystep) {$-1$};
 
  \fill[inactivecolor] (-1.5*\sc, 1*\ystep) circle (1.6pt);
  \fill[inactivecolor] (-1.0*\sc, 1*\ystep) circle (1.6pt);
  \fill[inactivecolor] ( 0.5*\sc, 1*\ystep) circle (1.6pt);
 
\end{scope}
 
\begin{scope}[xshift=15.5cm, local bounding box=panelC]
 
  \node[paneltitle] at (0, 2.0)
      {(c) Token sequence};
 
  \matrix (M) [matrix of math nodes, ampersand replacement=\&,
               nodes={inner sep=0pt, minimum width=15mm, minimum height=6mm,
                      text height=1.6ex, text depth=0.6ex,
                      font=\small, anchor=center,
                      draw=insetborder, line width=0.4pt},
               row 1/.style={nodes={fill=datacolor}},
               row 2/.style={nodes={fill=activecolor!20}},
               row 3/.style={nodes={fill=activecolor!20}},
               row 4/.style={nodes={fill=activecolor!20}},
               row sep=-0.4pt, column sep=-0.4pt] at (0, 0.4) {
    v \& \langle v, \mathbf{w}_i\rangle \& \Delta\chi(v, \mathbf{w}_i) \\
    v_1 \& -2.5 \& +1 \\
    v_5 \& \phantom{-}1.0 \& -1 \\
    v_6 \& \phantom{-}2.5 \& -1 \\
  };
 
\end{scope}
 
\end{tikzpicture}
}

\vspace{-8pt}
\caption{(a) Embedded simplicial complex $K$, consisting of $6$ vertices $v_1, \dots, v_6$, $8$ edges and $1$ face.
(b) Euler Characteristic Curve of $K$ in the direction $\mathbf{w}_i$. 
(c) Token sequence $\{(\langle v, \mathbf{w}_i \rangle, \Delta\chi(v, \mathbf{w}_i))\}_v$ after masking vertices without change.}
\label{fig:ecc}
\end{figure}
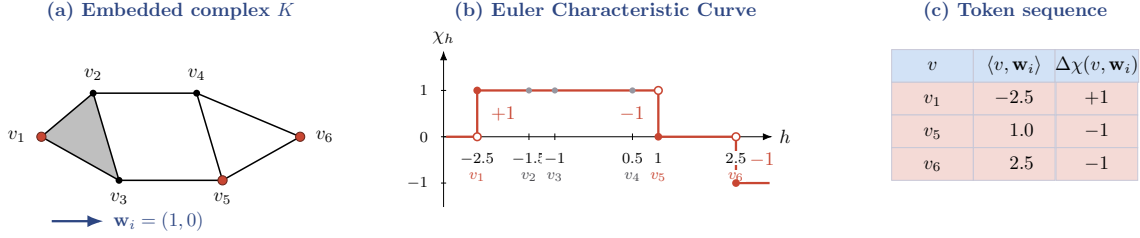

\begin{tcolorbox}
The continuous encoder removes a tuned resolution hyperparameter and the sigmoid approximation it requires, replacing the discretized ECT with a \(D \times d_{\mathrm{out}}\) feature matrix derived directly from the \(O(N)\) vertices of the complex. It substitutes for the discretized ECT, including in every representation architecture studied next.
\end{tcolorbox}

\section{ECT Representations} \label{sec:ect-representation}
The ECT has several notable equivariance properties. It is inherently equivariant with respect to permutation of the input nodes or points. It can be made translation-equivariant through centering the shape during preprocessing \citep{crawford2020predicting,munch2025invitation}. And it is equivariant under rotations of the embedded shape: rotating the input by an angle $\theta$ corresponds to a circular shift of the sampling directions by the same angle. In practice when computing the ECC at \(D\) regularly spaced directions, rotational equivariance holds for rotations by multiples of \(2\pi/D\). The preservation of these equivariances depends on the neural network architecture. 

We propose and evaluate six distinct neural network architectures for processing ECC encodings to generate ECT representations. The encoded ECT of a shape embedded in $\mathbb{R}^2$ can be represented as a matrix $\mathbf{E} \in \mathbb{R}^{D \times H}$, where $D$ is the number of sampling directions and $H$ is the number of height thresholds or the transformer output dimension. Each row corresponds to an ECC computed along a specific direction $\mathbf{w}_i$. 
All proposed architectures start with a continuous ECT layer followed by either discretization or an ECC transformer, producing an encoded ECT matrix. Their representation architecture then transforms the ECT to a (possibly equivariant) vector representation $\mathbf{z}$. And finally, a fully connected classification head is used to predict class probabilities. See also \cref{fig:pipeline}. 

The ECT is computed via a dedicated layer that processes embedded simplicial or cubical complexes. For fixed angles, the directions are regularly distributed according to $\theta_i = 2\pi i / D$ for $i \in \{0, 1, \ldots, D-1\}$. This regular distribution is essential for maintaining rotational equivariance, as discussed below. Alternatively, the direction corresponding to the angles can be optimized jointly with other network parameters to allow the network to learn task-specific directional sampling patterns,  at the cost of breaking rotational equivariance. 

\paragraph{Feedforward network.} The feedforward architecture takes the simplest approach by flattening the entire ECT matrix into a single vector and processing it through fully-connected layers. This baseline architecture makes no assumptions about the structure of the ECT and treats it as an arbitrary collection of features. Formally, the neural representation $\mathbf{z}$, from which we compute the output $y$ is just 
$\mathbf{z} = \phi(\mathrm{ECT})$ and $y = \psi(\mathbf{z})$, 
where the encoder $\phi$ extracts representation $\mathbf{z}$ from the ECT and the decoder $\psi$ produces the final output $y$. This architecture thus serves as a baseline that must learn rotation invariance purely from data rather than benefiting from architectural inductive biases. 

\paragraph{Deep Set.} Here we treat the ECT as an unordered set of ECCs, exploiting the permutation invariance property of set functions. Following the framework of \citet{zaheer2017deep}, the architecture consists of three components: a \emph{feature extraction network} $\phi$ applied independently to each ECC, a permutation-invariant \emph{aggregation function} $\oplus$, and a \emph{decoder} network $\psi$. 
For data in $\mathbb{R}^2$ with fixed, regularly distributed angles this architecture is only rotationally equivariant if the angle information is excluded as input features to $\phi$. However, excluding the angles as feature information is lossy, because there is no way to keep track of which angles are close to one another in the Deep Set architecture. In early experiments, this architecture did not learn well without angle information and we have therefore included angles in all Deep Sets experiments reported. 

\paragraph{2D convolution.}  The ECT matrix can be directly viewed as a 2D image, where one dimension corresponds to sampling directions and the other to height thresholds. This natural interpretation allows the application of standard 2D convolutions, with the network learning hierarchical spatial features through successive convolutional layers. The key to achieving rotational equivariance in this architecture lies in the careful implementation of padding. Specifically, rotational equivariance requires \emph{circular padding} along the directional dimension. When the input shape rotates by angle $\alpha$, the ECT matrix undergoes a corresponding circular shift along the direction axis. With circular padding, convolutions commute with these circular shifts, thereby preserving rotational equivariance. 
Note that the 2D convolution also has approximate equivariance to translations of the height thresholds, which is not a reasonable inductive bias, but we include the architecture because it has been suggested before \citep{mcguire2024classification, roell2024differentiable}.

\paragraph{1D convolution.} Instead, we apply 1D convolutions along the directional dimension. This approach can be understood as treating each row of the ECT matrix as a separate signal channel, with the convolutions capturing patterns in how the Euler characteristic varies across different sampling directions. As with the 2D convolutional case, rotational equivariance in the 1D architecture requires \emph{circular padding} along the directional dimension. 

\paragraph{1D complex convolution.} This architecture takes a fundamentally different approach by representing ECCs in the complex domain, where angular information is naturally encoded in the phase. This is inspired by rotationally equivariant learning from contours \citep{gardaa2025rotatouille}. Each encoded ECC at height $h$ is first converted to a complex representation by multiplying it with a phase factor determined by the sampling direction and then applying 1D complex convolution where all convolutions are in the complex domain and the activation function preserves the phase structure while applying nonlinearity to the magnitude. Complex convolutions in this form are inherently equivariant to rotations \citep{gardaa2025rotatouille}. 

\paragraph{Hybrid.} Combining the benefits of 1D convolutions with the flexibility of fully-connected layers, this architecture begins with 1D convolutional layers with circular padding. After these convolutional layers, the output is flattened and  processed by a fully-connected layer with ReLU activation. This design allows the network to first extract rotationally equivariant features through convolutions before using the greater flexibility of dense connections for final feature aggregation.

\begin{tcolorbox}
    The six architectures differ in their inductive biases about which structure of the ECT is informative: the feedforward network assumes none, the Deep Set assumes permutation symmetry of directions, the 2D/1D convolutions assume circular ordering of directions, and the complex convolution and hybrid assume rotational equivariance specifically. The next section measures which inductive biases pay off across point clouds, graphs, cubical complexes, and meshes.
\end{tcolorbox}

\section{Experiments} \label{sec:experiments}

The aim of our subsequent experiments is to analyze the ramifications of different ECC encoding and ECT representation strategies on downstream performance.

\subsection{Experimental Setup}
\begin{figure}
  \centering
  \resizebox{\textwidth}{!}{%
  \begin{tikzpicture}[node distance=0.4cm and 1.0cm]
    \node[datanode] (input) {\drawgraph\\[2pt]Input};
    \node[layernode, right=0.9cm of input] (centlayer) {Continuous ECT \\[-2pt]representation};

    \node[datanode,  right=1.6cm of centlayer, yshift=-1.6cm] (tokens)         {\drawtokens\\[2pt]Tokens};
    \node[layernode, right=0.7cm of tokens]                   (transformer_b)  {Transformer};

    \node[layernode] (disc_m)        at (tokens |- 0,0)        {Discretization};
    \node[layernode] (transformer_m) at (transformer_b |- 0,0) {Transformer};

    \node[layernode] (disc_t) at ($(tokens)!0.5!(transformer_b)$ |- centlayer) [yshift=3.2cm] {Discretization};

    \begin{scope}[on background layer]
      \node[groupbox, fit=(disc_t)(disc_m)(transformer_m)(tokens)(transformer_b),
            label={[font=\footnotesize\sffamily, color=bordercol]above:ECC encoder}]
            (encbox) {};
    \end{scope}

    \node[datanode, right=1.0cm of encbox] (ectdata) {\drawect\\[2pt]Encoded ECT};

    \node[choicenode, right=1.4cm of ectdata, yshift= 3.0cm] (ffn)    {Feedforward NN};
    \node[choicenode, right=1.4cm of ectdata, yshift= 1.8cm] (dset)   {Deep Set};
    \node[choicenode, right=1.4cm of ectdata, yshift= 0.6cm] (conv1d) {1D Convolution};
    \node[choicenode, right=1.4cm of ectdata, yshift=-0.6cm] (conv2d) {2D Convolution};
    \node[choicenode, right=1.4cm of ectdata, yshift=-1.8cm] (cconv)  {1D Complex Conv.};
    \node[choicenode, right=1.4cm of ectdata, yshift=-3.0cm] (hybrid) {Hybrid};
    \begin{scope}[on background layer]
      \node[groupbox, fit=(ffn)(dset)(conv1d)(conv2d)(cconv)(hybrid),
            label={[font=\footnotesize\sffamily, color=bordercol]above:Representation architecture}]
            (choicebox) {};
    \end{scope}

    \node[datanode, right=1.4cm of choicebox] (z)      {\drawzvec\\[2pt]$\mathbf{z}$};
    \node[layernode, right=0.9cm of z]        (head)   {Classification\\[-2pt]head};
    \node[datanode, right=0.9cm of head]      (output) {\drawyout\\[2pt]Output $\mathbf{y}$};

    \draw[arrow] (input)              -- (centlayer);
    \draw[arrow] (centlayer.east)     -- (disc_t.west);
    \draw[arrow] (centlayer.east)     -- (disc_m.west);
    \draw[arrow] (centlayer.east)     -- (tokens.west);
    \draw[arrow] (disc_m)             -- (transformer_m);
    \draw[arrow] (tokens)             -- (transformer_b);
    \draw[arrow] (disc_t.east)        -- (ectdata.west);
    \draw[arrow] (transformer_m.east) -- (ectdata.west);
    \draw[arrow] (transformer_b.east) -- (ectdata.west);
    \foreach \dest in {ffn, dset, conv1d, conv2d, cconv, hybrid}
      \draw[arrow] (ectdata.east) -- (\dest.west);
    \foreach \src in {ffn, dset, conv1d, conv2d, cconv, hybrid}
      \draw[arrow] (\src.east) -- (z.west);
    \draw[arrow] (z)    -- (head);
    \draw[arrow] (head) -- (output);
  \end{tikzpicture}}%
  \caption{Overview of the pipeline. The continuous ECT representation feeds three interchangeable paths inside the ECC encoder: discretization onto a fixed grid (top), discretization followed by a transformer (middle), or tokenization followed by a transformer (bottom). All three produce a $D \times H$ matrix that drives the same representation architecture, classification head, and output. Blue rectangles denote data; orange rectangles denote fixed layers; green rectangles denote choices.}
  \label{fig:pipeline}
\end{figure}
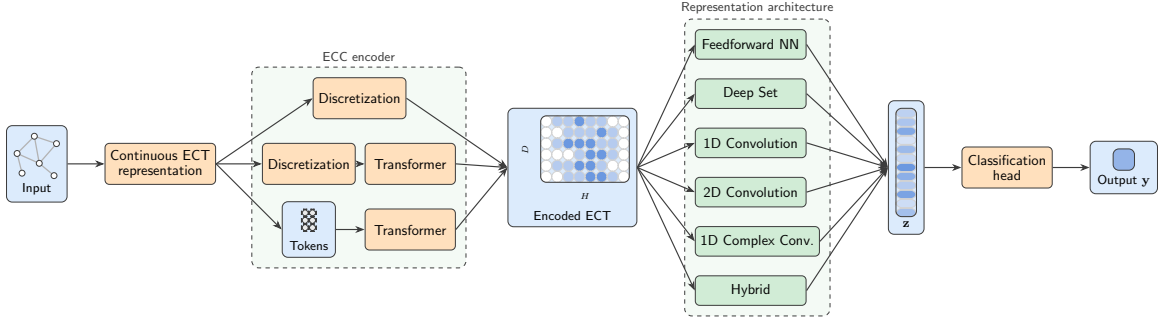

We implemented three ECC encoding strategies (discrete, discrete with a transformer applied to the discretized ECC and continuous) and all six ECT representation architectures described above and evaluated them on a range of classification tasks (\cref{fig:pipeline}). In the discrete case we used $H=32$ height thresholds, and in the continuous case the output dimension $d_{\mathrm{out}} = 32$ matches the resolution parameter used by the discretized ECT, ensuring identical downstream architectures for both variants. The ECT was computed using $D=64$ fixed directions, resulting in a $64 \times 32$ matrix representation for each shape. 
All models were trained using the Adam optimizer with standard PyTorch default hyperparameters ($\beta_1=0.9, \beta_2=0.999$), and cross-entropy loss for classification. For all experiments, we used a learning rate of $0.0001$ and whenever possible batch size of 128. Due to memory constraints, we reduced the batch size to 16 for the 3D mesh data. The main experiments are reported with fixed angles, with a separate experiment about learned angles for the point cloud dataset. Specific architectures are reported in the appendix. Our goal is to compare different neural representations of the ECT and do not include any architectures that bypass the ECT in our experiments. 

All architectures produce a 64-dimensional representation $\mathbf{z}$ that feeds into a common lightweight classification head. This head consists of two fully-connected layers: the first reduces from 64 to 16 dimensions with ReLU activation, and the second maps from 16 dimensions to the number of output classes with softmax activation.

We evaluate on four data types. Three are embedded in \(\mathbb{R}^2\): point clouds and cubical complexes derived from Fashion-MNIST~\citep{xiao2017fashion}, and the letter-low, letter-med, and letter-high graph datasets from TUDataset~\citep{morris2020tudataset}, which carry 2-dimensional vertex coordinates. The fourth is a triangle-mesh dataset in $\mathbb{R}^3$, SwissBuildings~\citep{swissbuildings3d}, a building-type classification task. For the 2D datasets, we sample $D=64$ directions equally spaced on the circle; for the 3D meshes, we sample them on a cylinder grid that preserves the circular structure along the azimuthal axis, so the convolutional architectures still apply circular padding there. See \cref{app:data} for more details on the SwissBuildings dataset.
All datasets were split into training and test data and for each of the 5 training runs, the training data was split into training and validation data. Because of their respective sizes, we trained the point cloud and mesh datasets for 20 epochs, whereas we trained the graph datasets for 100 epochs. We report the test accuracy for the epoch that achieved the best validation accuracy.
All code used to produce our results is available on \url{\repo}.

\subsection{Experimental Results}

\Cref{tab:res-main} presents the \emph{improvements} in accuracy of continuous ECC encoding compared to the discrete encoding. To distinguish the contributions of the tokenization and the transformer architecture, we also show the discrete version with additional transformer encoding. For classification accuracies for all encoders and representations across all benchmark datasets, see \cref{tab:res-complete} and \cref{fig:heatmaps}. The datasets span a wide difficulty range. Letter-low is nearly saturated, with all architectures exceeding $0.9$ accuracy. Note also that the Fashion-MNIST-derived datasets (FashionMNISTPointCloud and FashionMNISTCubical) and Letters datasets are not rotationally symmetric: the source images have a canonical upright orientation, so rotational equivariance is not a useful inductive bias for these tasks, which partly explains the strong performance of the feedforward network there.

\begin{table}
\centering

\sisetup{
    detect-all,
    table-format  = +1.3,
    print-mantissa-implicit-plus,
}

\begin{tabular}{lccc}
\toprule
Dataset & Discrete & Continuous & Discrete Transformer \\
\midrule
FashionMNISTPointCloud & 0.702 & \bfseries 0.778 (+0.077) & 0.717 (+0.016) \\
Letter-low & 0.936 & 0.959 (+0.023) & \bfseries 0.968 (+0.032) \\
Letter-med & 0.828 & \bfseries 0.843 (+0.015) & 0.820 (-0.008) \\
Letter-high & 0.751 & \bfseries 0.810 (+0.059) & 0.675 (-0.076) \\
SwissBuildings & 0.717 & \bfseries 0.812 (+0.095) & 0.680 (-0.037) \\
FashionMNISTCubical & 0.778 & \bfseries 0.837 (+0.059) & 0.768 (-0.010) \\
\bottomrule
\end{tabular}

\caption{For each dataset, the accuracy and in parenthesis the accuracy difference between the best discrete model and (i) the best continuous model and (ii) the best discrete-transformer model. Positive signs indicate that the presented versions improved accuracy, while negative signs indicate that the discrete version performed better.}
\label{tab:res-main}
\end{table}

The continuous encoding improves accuracy on all six datasets. The gains are largest on SwissBuildings $(+0.095)$ and FashionMNISTPointCloud $(+0.077)$, with smaller improvements on Letter-high $(+0.059)$ and FashionMNISTCubical $(+0.059)$. Adding a transformer on top of the discrete encoding produces no consistent gains, but for some of the representations, the transformer can have an effect (SwissBuildings accuracy in the feedforward architecture moves from $0.505$ to $0.606$). However, for the best architectures for each dataset, the benefit of the continuous representation comes from the tokenization itself rather than from the added transformer capacity. 

Overall, the influence of the ECT representation is smaller and less clear than the influence of the ECC encoder, but the two interact. With continuous encoding, the feedforward network is the strongest performer on four of the six datasets, edged out only on Letter-low by the hybrid 1D convolution and on SwissBuildings by the 1D complex convolution. With discrete encoding the feedforward network is unstable---standard errors above $0.2$ on FashionMNISTPointCloud and a collapse to an accuracy of $0.505$ on SwissBuildings---and the hybrid 1D convolution becomes the most reliable performer. Continuous tokenization narrows this gap by giving the feedforward network a more learnable input; the convolutional architectures, with their built-in directional structure, deliver more stable training across all three encoders.
\Cref{fig:datasize} compares sample efficiency on FashionMNISTPointCloud across architectures and encoders. The dominant effect is the positive interaction between the continuous encoder and the feedforward architecture. Only the hybrid architecture with any encoder, comes close to the feedforward architecture with the continuous encoder. The Deep Set and convolution architectures lag behind the top two performers throughout, regardless of encoder.

\section{Conclusion and Discussion}
We introduced a continuous tokenization of the ECT that records the net Euler-characteristic change attributed to each vertex along each direction, eliminating the height-resolution hyperparameter of the discretized ECT. Across point clouds, graphs, cubical complexes, and meshes, this representation improves accuracy on all of the six primary benchmarks, with gains up to $+0.095$ on SwissBuildings. The improvement is attributable to the tokenization itself rather than to added model capacity: applying a transformer on top of the discretized ECT does not give consistent gains. Among the representation architectures, the feedforward network is the strongest performer when paired with the continuous encoder for most datasets.

A natural direction for future work is extending the ECT representation framework to 3D shapes that are equivariant to general 3D rotations. The architectures presented here exploit the circular structure of the 2D rotations via circular padding and complex phase encoding. In 3D, the ECT instead integrates over directions on the sphere $S^2$ rather than $S^1$, requiring $SO(3)$-equivariance rather than $SO(2)$-equivariance. Several equivariant architectures have been developed for signals defined on the sphere \citep{esteves2020spinweighted, shen2020quaternion, ballerin2025so3equivariant}. Adapting the ECT representation architectures to the 3D setting will likely require borrowing from these approaches, for instance by treating the ECT as a scalar field on $S^2$ and applying spherical convolutions that commute with $SO(3)$ rotations. This is a non-trivial extension, as the efficient circular padding tricks used here have no direct spherical analogue and $SO(3)$-equivariant architectures are computationally demanding. But such extensions to the 3-dimensional case would substantially broaden the applicability of learned ECT representations to problems in 3D shape analysis.

\ifarXiv
    \acks{This work has received funding from the Swiss National Science Foundation (Grant 235713), the European Union (Grant 101126560) and the Swiss State Secretariat for Education, Research, and Innovation~(SERI).}
\fi

\bibliography{refs}

@article{turner2014persistent,
  title={Persistent homology transform for modeling shapes and surfaces},
  author={Turner, Katharine and Mukherjee, Sayan and Boyer, Doug M},
  journal={Information and Inference: A Journal of the IMA},
  volume={3},
  number={4},
  pages={310--344},
  year={2014},
  publisher={Oxford University Press}
}

@Article{Ghrist2018,
    author="Ghrist, Robert and Levanger, Rachel and Mai, Huy",
    title="Persistent homology and Euler integral transforms",
    journal="Journal of Applied and Computational Topology",
    year="2018",
    volume="2",
    number="1",
    pages="55--60",
    issn="2367-1734",
    doi="10.1007/s41468-018-0017-1",
    url="https://doi.org/10.1007/s41468-018-0017-1"
}

@article{curry2022many,
  title={How many directions determine a shape and other sufficiency results for two topological transforms},
  author={Curry, Justin and Mukherjee, Sayan and Turner, Katharine},
  journal={Transactions of the American Mathematical Society, Series B},
  volume={9},
  number={32},
  pages={1006--1043},
  year={2022}
}

@inproceedings{roell2024differentiable,
	author = {Ernst R{\"o}ell and Bastian Rieck},
	booktitle = {International Conference on Learning Representations},
	title = {Differentiable {E}uler Characteristic Transforms for Shape Classification},
	url = {https://openreview.net/forum?id=MO632iPq3I},
	year = {2024},
}

@inproceedings{roell2026ipt,
  title         = {Point Cloud Synthesis Using Inner Product Transforms},
  author        = {Ernst R{\"o}ell and Bastian Rieck},
  author+an     = {2=highlight},
  year          = 2025,
  booktitle     = {Advances in Neural Information Processing Systems},
  volume        = {38},
  pages         = {14551--14583},
  publisher     = {Curran Associates, Inc.},
  editor        = {D. Belgrave and C. Zhang and H. Lin and R. Pascanu and P. Koniusz and M. Ghassemi and N. Chen},
}

@article{zaheer2017deep,
  title={Deep sets},
  author={Zaheer, Manzil and Kottur, Satwik and Ravanbakhsh, Siamak and Poczos, Barnabas and Salakhutdinov, Russ R and Smola, Alexander J},
  journal={Advances in neural information processing systems},
  volume={30},
  year={2017}
}

@misc{mcguire2024classification,
  title={Classification of {E}uler {C}haracteristic {T}ransforms using {C}onvolutional {N}eural {N}etworks},
  author={McGuire, Sarah},
  howpublished={Applied Algebraic Topology Network Talk},
  month={May},
  year={2024}, 
  url={https://www.youtube.com/watch?v=haijmIm6K8o}
}

@article{crawford2020predicting,
  title={Predicting clinical outcomes in glioblastoma: an application of topological and functional data analysis},
  author={Crawford, Lorin and Monod, Anthea and Chen, Andrew X and Mukherjee, Sayan and Rabad{\'a}n, Ra{\'u}l},
  journal={Journal of the American Statistical Association},
  volume={115},
  number={531},
  pages={1139--1150},
  year={2020},
  publisher={Taylor \& Francis}
}

@article{munch2025invitation,
  title={An invitation to the {E}uler characteristic transform},
  author={Munch, Elizabeth},
  journal={The American Mathematical Monthly},
  volume={132},
  number={1},
  pages={15--25},
  year={2025},
  publisher={Taylor \& Francis}
}

@inproceedings{nadimpalli2023euler,
  title={Euler characteristic transform based topological loss for reconstructing 3D images from single 2D slices},
  author={Nadimpalli, Kalyan Varma and Chattopadhyay, Amit and Rieck, Bastian},
  booktitle={Proceedings of the IEEE/CVF Conference on Computer Vision and Pattern Recognition},
  pages={571--579},
  year={2023}
}

@inproceedings{rohrscheidt2025disslect,
  title         = {{Diss-l-ECT}: Dissecting Graph Data with Local {E}uler Characteristic Transforms},
  author        = {von Rohrscheidt, Julius and Rieck, Bastian},
  booktitle     = {Proceedings of the 42nd International Conference on Machine Learning},
  series        = {Proceedings of Machine Learning Research},
  publisher     = {PMLR},
  volume        = 267,
  pages         = {61790--61809},
  editor        = {Singh, Aarti and Fazel, Maryam and Hsu, Daniel and Lacoste-Julien, Simon and Berkenkamp, Felix and Maharaj, Tegan and Wagstaff, Kiri and Zhu, Jerry},
  year          = 2025,
  eprint        = {2410.02622},
  archiveprefix = {arXiv},
  primaryclass  = {cs.LG},
}

@inproceedings{amboage2026leap,
    title={{LEAP}: Local {ECT}-Based Learnable Positional Encodings for Graphs},
    author={Juan P Garcia Amboage and Ernst R{\"o}ell and Patrick Schnider and Bastian Rieck},
    booktitle={International Conference on Learning Representations},
    year={2026},
    url={https://openreview.net/forum?id=8XFPhByERc}
}

@inproceedings{toscano2025molecular,
  title         = {Molecular Machine Learning Using {E}uler Characteristic Transforms},
  author        = {Victor Toscano-Duran and Florian Rottach and Bastian Rieck},
  author+an     = {3=highlight},
  year          = 2026,
  booktitle     = {Artificial Intelligence in Biomedicine},
  publisher     = {Springer},
  address       = {Cham, Switzerland},
  pages         = {391--405},
  editor        = {L{\'o}pez Fern{\'a}ndez, Aurelio and Rodr{\'i}guez-Gonz{\'a}lez, Alejandro and Leir{\'o}s-Rodr{\'i}guez, Raquel and Mata Miquel, Christian and Gonz{\'a}lez Su{\'a}rez, V{\'i}ctor Manuel},
  eprint        = {2507.03474},
  archiveprefix = {arXiv},
  primaryclass  = {cs.LG},
}

@inproceedings{gardaa2025rotatouille,
title={RotaTouille: Rotation Equivariant Deep Learning for Contours},
author={Odin Hoff Gardaa and Nello Blaser},
booktitle={The Fourth Learning on Graphs Conference},
year={2025},
url={https://openreview.net/forum?id=VbCLyz3uW7}
}

@inproceedings{morris2020tudataset,
  title={TUDataset: A collection of benchmark datasets for learning with graphs},
  author={Morris, Christopher and Kriege, Nils Morten and Bause, Franka and Kersting, Kristian and Mutzel, Petra and Neumann, Marion},
  booktitle={ICML 2020 Workshop on Graph Representation Learning and Beyond},
  year={2020}
}

@misc{xiao2017fashion,
  author       = {Han Xiao and Kashif Rasul and Roland Vollgraf},
  title        = {Fashion-{MNIST}: a Novel Image Dataset for Benchmarking Machine Learning Algorithms},
  date         = {2017-08-28},
  year         = {2017},
  eprintclass  = {cs.LG},
  eprinttype   = {arXiv},
  eprint       = {cs.LG/1708.07747},
}

@inproceedings{esteves2020spinweighted,
 author = {Esteves, Carlos and Makadia, Ameesh and Daniilidis, Kostas},
 booktitle = {Advances in Neural Information Processing Systems},
 editor = {H. Larochelle and M. Ranzato and R. Hadsell and M.F. Balcan and H. Lin},
 pages = {8614--8625},
 publisher = {Curran Associates, Inc.},
 title = {Spin-Weighted Spherical {CNNs}},
 volume = {33},
 year = {2020}
}

@InProceedings{shen2020quaternion,
    author="Shen, Wen and Zhang, Binbin and Huang, Shikun and Wei, Zhihua and Zhang, Quanshi",
    editor="Vedaldi, Andrea and Bischof, Horst and Brox, Thomas and Frahm, Jan-Michael",
    title="3D-Rotation-Equivariant Quaternion Neural Networks",
    booktitle="Computer Vision -- ECCV 2020",
    year="2020",
    publisher="Springer International Publishing",
    address="Cham",
    pages="531--547",
}

@misc{ballerin2025so3equivariant,
      title={{SO(3)}-Equivariant Neural Networks for Learning from Scalar and Vector Fields on Spheres}, 
      author={Francesco Ballerin and Nello Blaser and Erlend Grong},
      year={2026},
      eprint={2503.09456},
      archivePrefix={arXiv},
      primaryClass={cs.LG},
      url={https://arxiv.org/abs/2503.09456}, 
}

@misc{swissbuildings3d,
  author       = {{Federal Office of Topography swisstopo}},
  title        = {{swissBUILDINGS3D 3.0 Beta}},
  year         = {2025},
  howpublished = {\url{https://www.swisstopo.admin.ch/en/landscape-model-swissbuildings3d-3-0-beta}},
  note         = {Accessed 29 April 2026.},
}

@incollection {MR1448182,
    AUTHOR = {Schapira, P.},
     TITLE = {Tomography of constructible functions},
 BOOKTITLE = {Applied algebra, algebraic algorithms and error-correcting codes ({P}aris, 1995)},
    SERIES = {Lecture Notes in Comput. Sci.},
    VOLUME = {948},
     PAGES = {427--435},
 PUBLISHER = {Springer, Berlin},
      YEAR = {1995},
      ISBN = {3-540-60114-7},
   MRCLASS = {32L25 (44A12 92C55)},
  MRNUMBER = {1448182},
       URL = {https://doi.org/10.1007/3-540-60114-7_33},
}

\clearpage
\let\cleardoublepage\clearpage
\appendix

\counterwithin*{figure}{part}
\counterwithin*{table}{part}
\stepcounter{part}
\renewcommand{\thefigure}{S.\arabic{figure}}
\renewcommand{\thetable}{S.\arabic{table}}

\section{Architecture Specifications} \label{app:specs}
Below we list detailed descriptions of each architecture as used in the experiments. \Cref{tab:params} shows the number of parameters for the different representation models. The classification heads are excluded because their number of parameters depends on the number of classes in the different datasets. 
Note that the transformer architecture for the continuous ECT has 69216 parameters and roughly doubles the total number of parameters. Similarly, the transformer architecture for the discrete ECT has 69152 parameters. 
\begin{table}[ht]
    \centering
\begin{tabular}{lr}
\toprule
Model & Parameters \\
\midrule
Hybrid 1D convolution & 85264 \\
1D convolution & 78272 \\
2D convolution & 81840 \\
1D complex convolution & 96256 \\
Feedforward network & 131136 \\
Deep Set & 78528 \\
\bottomrule
\end{tabular}
\caption{Number of parameters for the different models. Complex parameters are counted as 2 real parameters.}
\label{tab:params}
\end{table}

The encoded ECT is a $D \times H$ matrix ($D = 64$ directions, $H = 32$ heights or transformer output dimension). The feedforward and 2D convolutional architectures consume it in this directions-first layout. The Conv1D-based architectures (Deep Set, 1D conv, 1D complex conv, hybrid) transpose it to channel-first form so the convolution slides along the directional axis $D$, which becomes the length, while the height/feature axis becomes the channels; when angle information is added as an extra channel the channel count is $H + 1$.

\subsection{Feedforward Network Architecture}

The Feedforward network provides a simple baseline by flattening the entire $D \times H = 64 \times 32$ ECT matrix into a 2048-dimensional vector and processing it through a single fully-connected layer that reduces to 64 dimensions, followed by ReLU activation. This architecture makes no structural assumptions about the ECT and does not incorporate angle information, treating the representation purely as an unordered collection of features. As discussed, this flattening operation breaks rotational equivariance by imposing a fixed ordering on the directional dimension.

\subsection{Deep Set Architecture}
Our Deep Set implementation uses four pointwise Conv1D layers with kernel size 1, effectively implementing the per-curve feature extraction network $\phi$ through 1D convolutions applied independently to each direction. The architecture progressively transforms the $(H+1)$-dimensional input (which includes one angle feature) through hidden dimensions of 128, 256, 128, and finally to 64 features, with ReLU activations between each layer. The aggregation function $\oplus$ is implemented as max pooling over the directional dimension, which combines the 64-dimensional representations from all $D=64$ directions into a single 64-dimensional feature vector. This vector then feeds into the shared classification head.

\subsection{2D Convolutional Architecture}
The 2D CNN processes the ECT as a two-channel image with dimensions $64 \times 32 \times 2$ (directions × heights × channels). The first channel contains the ECT values themselves, while the second channel contains angle differences broadcast across all height thresholds. The architecture consists of four convolutional layers with progressively increasing channel counts: the first layer transforms the 2-channel input to 16 features using a $3 \times 3$ kernel, followed by layers that increase the channel count to 32, maintain it at 32, and finally expand to 64 features. The kernel sizes also increase for the deeper layers, with the third and fourth layers using $5 \times 5$ kernels to capture larger spatial patterns. Each convolution is followed by ReLU activation, and after the final layer, global max pooling aggregates information across both spatial dimensions to produce a 64-dimensional feature vector for classification. An important implementation detail for maintaining rotational equivariance is the custom padding scheme, using circular padding on the directional dimension and standard zero padding on the height dimension. 

\subsection{1D Convolutional Architecture}
The 1D CNN  applies convolutions along the directional dimension, processing a transposed $(H+1) \times D = 33 \times 64$-dimensional input. The architecture employs four 1D convolutional layers with progressively increasing kernel sizes to capture patterns at multiple scales. Starting from the $(H+1)$-dimensional input (which includes angle information), the first layer with kernel size 1 expands to 128 channels, followed by a second layer with kernel size 3 that reduces to 64 channels. The third and fourth layers maintain 64 channels while using larger kernels of size 5 and 7 respectively, allowing the network to integrate information across increasingly large directional neighborhoods. All convolutions use circular padding to enable rotational equivariance, and ReLU activations follow each layer. After the final convolution, max pooling over the directional dimension produces a 64-dimensional feature representation.

\subsection{1D Complex Convolutional Architecture}
The complex CNN operates entirely in the complex domain. The architecture begins by converting the real-valued ECT to a complex polar representation where each value is multiplied by $e^{i\theta_i}$ determining the phase. This creates a complex-valued $H \times D = 32 \times 64$-dimensional input that naturally encodes both the ECT magnitudes and the directional angles. Four complex convolutional layers process this representation with circular padding. The first layer uses kernel size 1 to expand from $H=32$ input channels to 32 complex feature channels, while the second layer with kernel size 3 increases this to 64 channels. The third and fourth layers maintain 64 channels using kernel size 5, allowing the network to capture increasingly complex patterns in the complex-valued features. Between layers, we apply the ModTanh activation function, which extends the standard tanh nonlinearity to complex numbers by computing $\text{ModTanh}(z) = \tanh(|z|) \cdot z / (|z| + \epsilon)$ with $\epsilon = 10^{-6}$. This activation preserves the phase structure of the complex representations while applying nonlinearity to the magnitudes. After the final complex convolution, we convert back to real values by computing the magnitude $|z|$ of each complex feature. Max pooling over the directional dimension then aggregates these magnitude features to produce the final 64-dimensional representation. 

\subsection{Hybrid Architecture}
The hybrid begins with three 1D convolutional layers with circular padding and progressively increasing kernel sizes (1, 3, and 5), transforming the $(H+1) \times D$ input through channel dimensions of 64, 64, and finally 16. After these convolutional layers, the output is flattened to a $D \times 16 = 1024$ dimensional vector, which is then processed by a fully-connected layer that reduces to 64 dimensions with ReLU activation. 

\subsection{Equivariance}
All the stated equivariance properties have been experimentally confirmed. Note that when angles are learnable parameters rather than fixed and regularly distributed, the rotational equivariance of all architectures is fundamentally broken. This is of course desirable, because it only makes sense to learn angles if some angles contain more relevant signal than others, in which case the task is not rotationally equivariant. In these cases, we can incorporate angles into the architectures dependent on the model architecture. The feedforward network does not incorporate angle information at all because that would be the same for all datapoints and the Deep Set architecture always concatenates the angle as an additional feature to each ECC. The 1D, hybrid, and 2D convolutional architectures add the difference between adjacent angles as an additional channel. The 1D complex convolution architecture always includes angle information.

\subsection{Directional Transformer Encoder for Continuous and Discrete ECC}

The directional transformer was configured with the hyperparameters listed in \cref{tab:continuous-ect-hyperparams}. 

\begin{table}[ht]
    \centering
    \begin{tabular}{lc}
        \toprule
        Hyperparameter & Value \\
        \midrule
        Token dimension (input) & 2 \\
        Model width $d_{\mathrm{model}}$ & 64 \\
        Attention heads & 4 \\
        Encoder layers & 2 \\
        Feedforward width & 128 \\
        Output dimension $d_{\mathrm{out}}$ & 32 \\
        \bottomrule
    \end{tabular}
    \caption{Hyperparameters of the directional transformer encoder.}
    \label{tab:continuous-ect-hyperparams}
\end{table}

\section{Dataset Construction} \label{app:data}

\paragraph{Fashion-MNIST point cloud.} 
To create the point cloud data, each image was converted by treating pixel intensities as a probability distribution and sampling 200 planar coordinates via multinomial sampling with replacement. The integer coordinates were then perturbed with uniform noise in the range $[-0.5, 0.5]$ to produce continuous-valued point clouds.

\paragraph{Fashion-MNIST cubical complex.}
Each grayscale image was binarized at a fixed intensity threshold $0.3$, and the resulting foreground pixels were used to construct a 2D cubical complex with vertices, edges between horizontally or vertically adjacent pixels, and square 2-cells for each filled $2 \times 2$ block. 

\paragraph{SwissBuildings.}
Derived from the \texttt{swissBUILDINGS3D 3.0 Beta} release, this building-classification dataset has labels indicating building type. We restricted the dataset to include the 10 classes with more than 1000 buildings. To avoid geographic leakage between training and test data, we split by geographic tile rather than by individual building. The class distribution is heavily imbalanced, so we apply per-epoch class-balanced subsampling: at the start of every epoch, each class is capped at the size of the smallest class by drawing a random subsample of indices.
For this dataset we sampled directions on a cylinder grid: $n_\phi$ azimuthal angles equally spaced on $[0, 2\pi)$ paired with $n_z$ elevations equally spaced on $[-1, 1]$, and then project the resulting $(n_\phi \cdot n_z)$ points onto the sphere by normalization. This preserves the circular structure along the $\phi$ axis, so the convolutional architectures still apply circular padding there. We use $D = n_\phi \cdot n_z = 64$ with $n_z = 4$. 

\section{Complete results} \label{app:res}
\begin{figure}
    \centering
    \includegraphics[width=\linewidth]{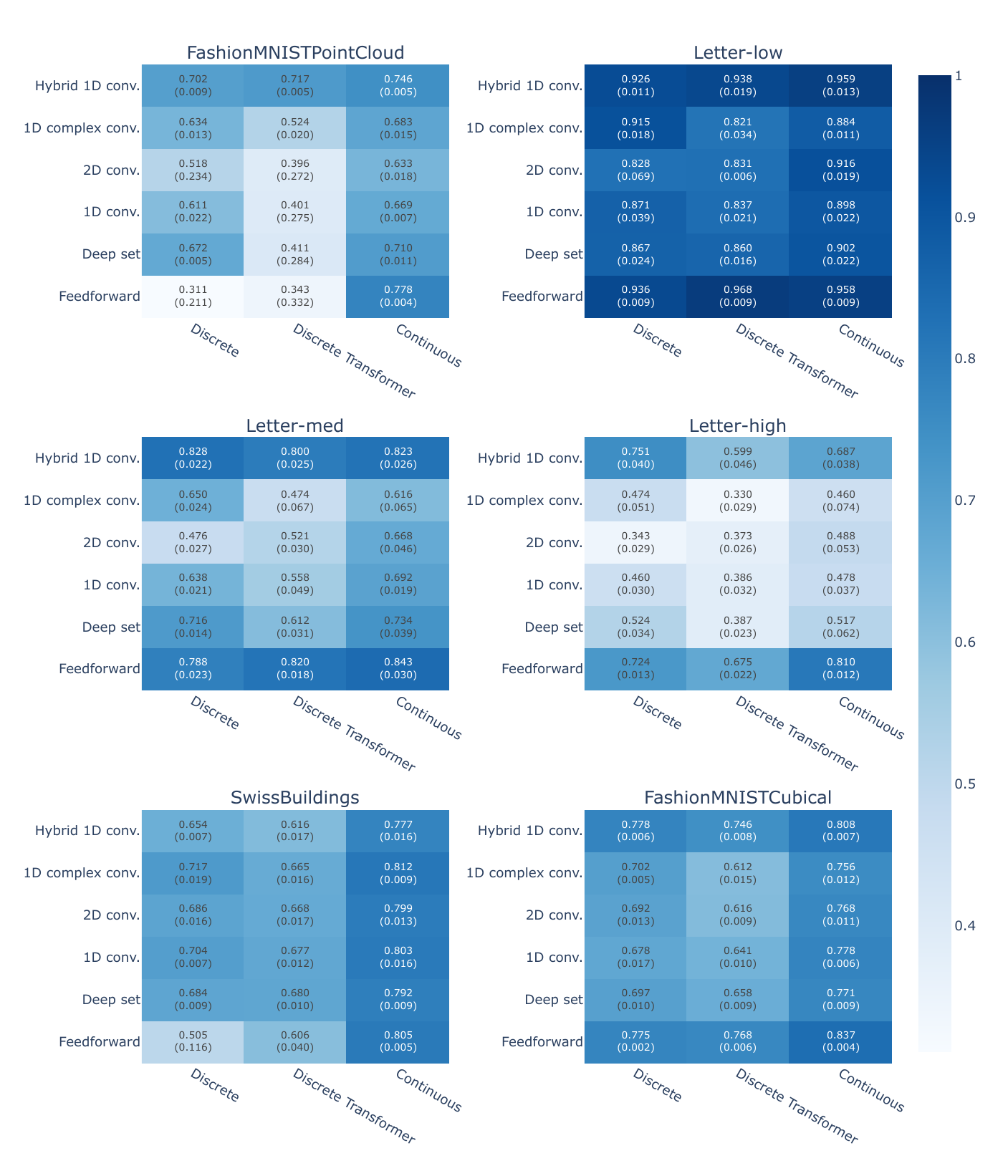}
    \caption{Test accuracies for the continuous, discrete, and discrete transformed ECT for all datasets. Standard errors (in parenthesis) are computed from 5 repeated runs.}
    \label{fig:heatmaps}
\end{figure}

\begin{figure}
    \centering
    \includegraphics[width=\textwidth]{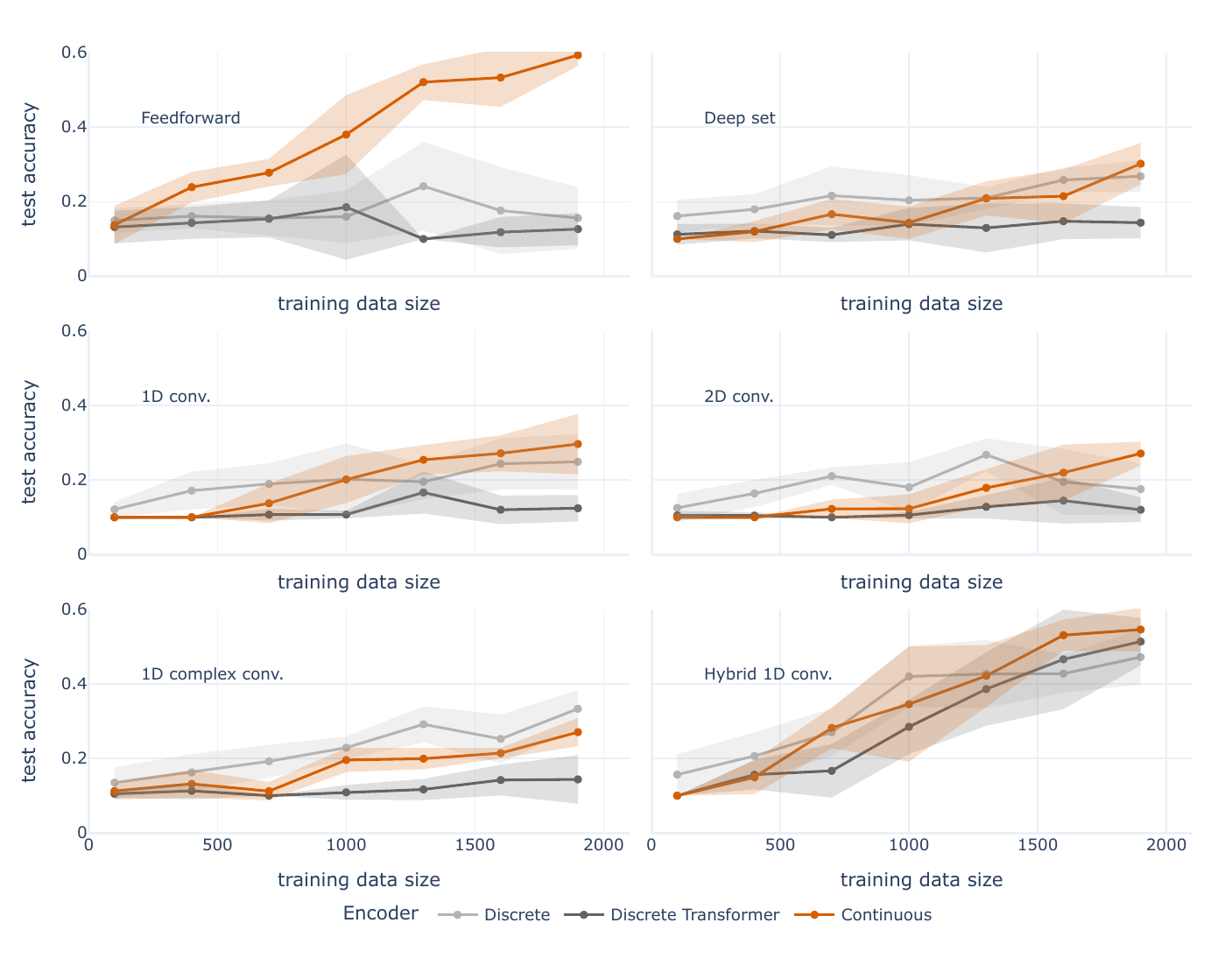}
    \caption{Test accuracy on  FashionMNISTPointCloud for different models for increasing training data sizes. Error bands are standard errors computed from 5 repeated runs.}
    \label{fig:datasize}
\end{figure}

\cref{tab:res-complete} and \cref{fig:heatmaps} show the complete results of test set accuracies for all tested models and across all datasets. \cref{fig:datasize} shows how test accuracy on the FashionMNIST-PointCloud data increases with training data size. 
\begin{table}
\centering
\resizebox{\textwidth}{!}{

\begin{tabular}{llcccccc}
\toprule
Encoder & Representation & FashionMNIST- & Letter- & Letter- & Letter- & SwissBuildings & FashionMNIST- \\
        &                & PointCloud    & low     & med     & high    &                & Cubical       \\
\midrule
\multirow[t]{6}{*}{Continuous} & 1D complex conv. & 0.683 (0.015) & 0.884 (0.011) & 0.616 (0.065) & 0.460 (0.074) & \bfseries \textcolor{cyan}{0.812 (0.009)} & 0.756 (0.012) \\
 & 1D conv. & 0.669 (0.007) & 0.898 (0.022) & 0.692 (0.019) & 0.478 (0.037) & \bfseries \textcolor{cyan}{0.803 (0.016)} & 0.778 (0.006) \\
 & 2D conv. & 0.633 (0.018) & 0.916 (0.019) & 0.668 (0.046) & 0.488 (0.053) & 0.799 (0.013) & 0.768 (0.011) \\
 & Deep set & 0.710 (0.011) & 0.902 (0.022) & 0.734 (0.039) & 0.517 (0.062) & 0.792 (0.009) & 0.771 (0.009) \\
 & Feedforward & \bfseries \textcolor{cyan}{0.778 (0.004)} & \textcolor{cyan}{0.958 (0.009)} & \bfseries \textcolor{cyan}{0.843 (0.030)} & \bfseries \textcolor{cyan}{0.810 (0.012)} & \bfseries \textcolor{cyan}{0.805 (0.005)} & \bfseries \textcolor{cyan}{0.837 (0.004)} \\
 & Hybrid 1D conv. & 0.746 (0.005) & \bfseries \textcolor{cyan}{0.959 (0.013)} & \bfseries \textcolor{cyan}{0.823 (0.026)} & 0.687 (0.038) & 0.777 (0.016) & 0.808 (0.007) \\
\multirow[t]{6}{*}{Discrete} & 1D complex conv. & 0.634 (0.013) & 0.915 (0.018) & 0.650 (0.024) & 0.474 (0.051) & \textcolor{cyan}{0.717 (0.019)} & 0.702 (0.005) \\
 & 1D conv. & 0.611 (0.022) & 0.871 (0.039) & 0.638 (0.021) & 0.460 (0.030) & \textcolor{cyan}{0.704 (0.007)} & 0.678 (0.017) \\
 & 2D conv. & 0.518 (0.234) & 0.828 (0.069) & 0.476 (0.027) & 0.343 (0.029) & 0.686 (0.016) & 0.692 (0.013) \\
 & Deep set & 0.672 (0.005) & 0.867 (0.024) & 0.716 (0.014) & 0.524 (0.034) & 0.684 (0.009) & 0.697 (0.010) \\
 & Feedforward & 0.311 (0.211) & \textcolor{cyan}{0.936 (0.009)} & 0.788 (0.023) & \textcolor{cyan}{0.724 (0.013)} & 0.505 (0.116) & \textcolor{cyan}{0.775 (0.002)} \\
 & Hybrid 1D conv. & \textcolor{cyan}{0.702 (0.009)} & 0.926 (0.011) & \bfseries \textcolor{cyan}{0.828 (0.022)} & \textcolor{cyan}{0.751 (0.040)} & 0.654 (0.007) & \textcolor{cyan}{0.778 (0.006)} \\
\multirow[t]{6}{*}{\makecell[lt]{Discrete\\Transformer}} & 1D complex conv. & 0.524 (0.020) & 0.821 (0.034) & 0.474 (0.067) & 0.330 (0.029) & 0.665 (0.016) & 0.612 (0.015) \\
 & 1D conv. & 0.401 (0.275) & 0.837 (0.021) & 0.558 (0.049) & 0.386 (0.032) & \textcolor{cyan}{0.677 (0.012)} & 0.641 (0.010) \\
 & 2D conv. & 0.396 (0.272) & 0.831 (0.006) & 0.521 (0.030) & 0.373 (0.026) & 0.668 (0.017) & 0.616 (0.009) \\
 & Deep set & 0.411 (0.284) & 0.860 (0.016) & 0.612 (0.031) & 0.387 (0.023) & \textcolor{cyan}{0.680 (0.010)} & 0.658 (0.009) \\
 & Feedforward & 0.343 (0.332) & \bfseries \textcolor{cyan}{0.968 (0.009)} & \bfseries \textcolor{cyan}{0.820 (0.018)} & \textcolor{cyan}{0.675 (0.022)} & 0.606 (0.040) & \textcolor{cyan}{0.768 (0.006)} \\
 & Hybrid 1D conv. & \textcolor{cyan}{0.717 (0.005)} & 0.938 (0.019) & 0.800 (0.025) & 0.599 (0.046) & 0.616 (0.017) & 0.746 (0.008) \\
\bottomrule
\end{tabular}

}
\caption{Test accuracies for the continuous, discrete, and discrete transformed ECT. Standard errors are computed from 5 repeated runs. Best results per encoder are colored, best overall results are bold. Multiple results are highlighted if within one standard error of the maximal value.}
\label{tab:res-complete}
\end{table}
%
\end{document}